\documentclass[letterpaper, 10 pt, conference]{ieeeconf}  % Comment this line out if you need a4paper

\IEEEoverridecommandlockouts                              % This command is only needed if 
\usepackage{graphics} % for pdf, bitmapped graphics files
\usepackage{epsfig} % for postscript graphics files
\usepackage{mathptmx} % assumes new font selection scheme installed
\usepackage{times} % assumes new font selection scheme installed
\usepackage{amsmath} % assumes amsmath package installed
\usepackage{amssymb}  % assumes amsmath package installed
\usepackage{xspace}
\newcommand{\sysname}{\texttt{HyperDet}\xspace}
\usepackage{booktabs}
\usepackage{xcolor}
\usepackage{booktabs}
\usepackage{multirow}
\usepackage{algorithm}
\usepackage{algorithmic}
\usepackage{subcaption}

\title{\LARGE \bf
HyperDet: 3D Object Detection with Hyper 4D Radar Point Clouds
}

\author{
  Yichun Xiao$^{1,2}$, Jin Jin$^{3}$, Runwei Guan$^{1}$,  
  {Fangqiang Ding}$^{1,*}$\thanks{*Corresponding author. Email: \texttt{fangqiangd@hkust-gz.edu.cn}} \\
\\
  $^1$HKUST(GZ)~
  $^2$University of Edinburgh~ 
  $^3$University of Oxford~
}

\begin{document}

\maketitle
\thispagestyle{empty}
\pagestyle{empty}

%%%%%%%%%%%%%%%%%%%%%%%%%%%%%%%%%%%%%%%%%%%%%%%%%%%%%%%%%%%%%%%%%%%%%%%%%%%%%%%%

\begin{abstract}
How far can 3D object detection go using 4D radar alone? Despite offering weather-robust and velocity-aware sensing for autonomous perception, modern 4D radar still yields sparse, noisy, and unstable point clouds, limiting radar-only 3D detection. We present \sysname, a detector-agnostic input enhancement pipeline that constructs task-aware hyper 4D radar point clouds by combining measured observations with completed foreground geometry. \sysname first refines short-window surround-view radar observations through spatio-temporal accumulation and cross-sensor validation, while Doppler-guided motion compensation reduces dynamic object trails when motion can be estimated reliably. It then performs foreground generative enhancement using LiDAR-guided pseudo-radar supervision available only during training, enriching object geometry while preserving measured radar background and radar-native attributes. During detector training, radar-aware object-level augmentation maintains Doppler consistency under geometric relocation. At inference, \sysname requires radar input alone and can be directly paired with standard 3D detectors. Experiments on two public surround-view 4D radar datasets demonstrate consistent improvements over matched temporal accumulation across standard 3D detectors, validating input-level radar enhancement as an effective approach to radar-only 3D detection.
\end{abstract}
\section{Introduction}

Reliable 3D object detection is fundamental to autonomous robots and vehicles operating in open and dynamic environments, providing essential scene understanding for downstream tracking, motion forecasting, and planning. Cameras and LiDAR underpin modern perception systems, yet their reliability can deteriorate in adverse weather and low visibility conditions~\cite{Karangwa2023review, Resilient24Chong}. 4D mmWave radar offers a complementary sensing profile. It directly measures radial velocity, is comparatively robust to weather and illumination variation, and can be deployed at relatively low cost~\cite{venon2022survey,ding2023hidden,ding2024radarocc}. These properties make radar particularly appealing for reliable perception in challenging operating conditions. However, radar-only 3D detection continues to lag substantially behind LiDAR-based methods on established benchmarks~\cite{palffy2022VoD}. A key bottleneck is the radar representation itself. Despite advances in MIMO-based radar design that improve angular resolution and enable elevation-aware sensing~\cite{jiang2024resolution}, 4D radar point clouds remain sparse, irregular, and noisy, yielding incomplete and unstable geometric evidence for accurate 3D bounding box estimation~\cite{venon2022survey}.

Current radar-only 3D detection research builds on LiDAR-inspired architectures, adapting them to radar point clouds through radar-specific feature encoders and network designs~\cite{zhou2023rmsaNet,musiat2024radarpillars,li2023pillarDan}. These approaches improve the encoding and processing of radar measurements, but do not directly remedy the sparse, noisy, and unstable geometry presented to the detector, which must still infer object location, extent, and orientation from limited geometric evidence. Improving the radar representation before detection is therefore a natural complementary direction, yet it cannot be achieved through naive densification alone.
Temporal accumulation increases point support but also produces moving-object trails that distort object extent and orientation~\cite{tan20223radardetection,palmer2023ego}. Aggregating returns from multiple surrounding radars expands spatial coverage, yet may also accumulate clutter and view-dependent inconsistencies. Even after spatio-temporal aggregation, measured returns remain too sparse to describe complete foreground geometry.
Radar measurements further impose constraints during detector training: LiDAR-style object-level copy-paste alters the line-of-sight direction of relocated returns, rendering their original radial velocities physically inconsistent with the new geometry. Together, these challenges call for a radar-aware input representation that strengthens object-level geometry and temporal coherence while preserving radar-native measurement cues.

\begin{figure}[t]
    \centering
    \includegraphics[width=\linewidth]{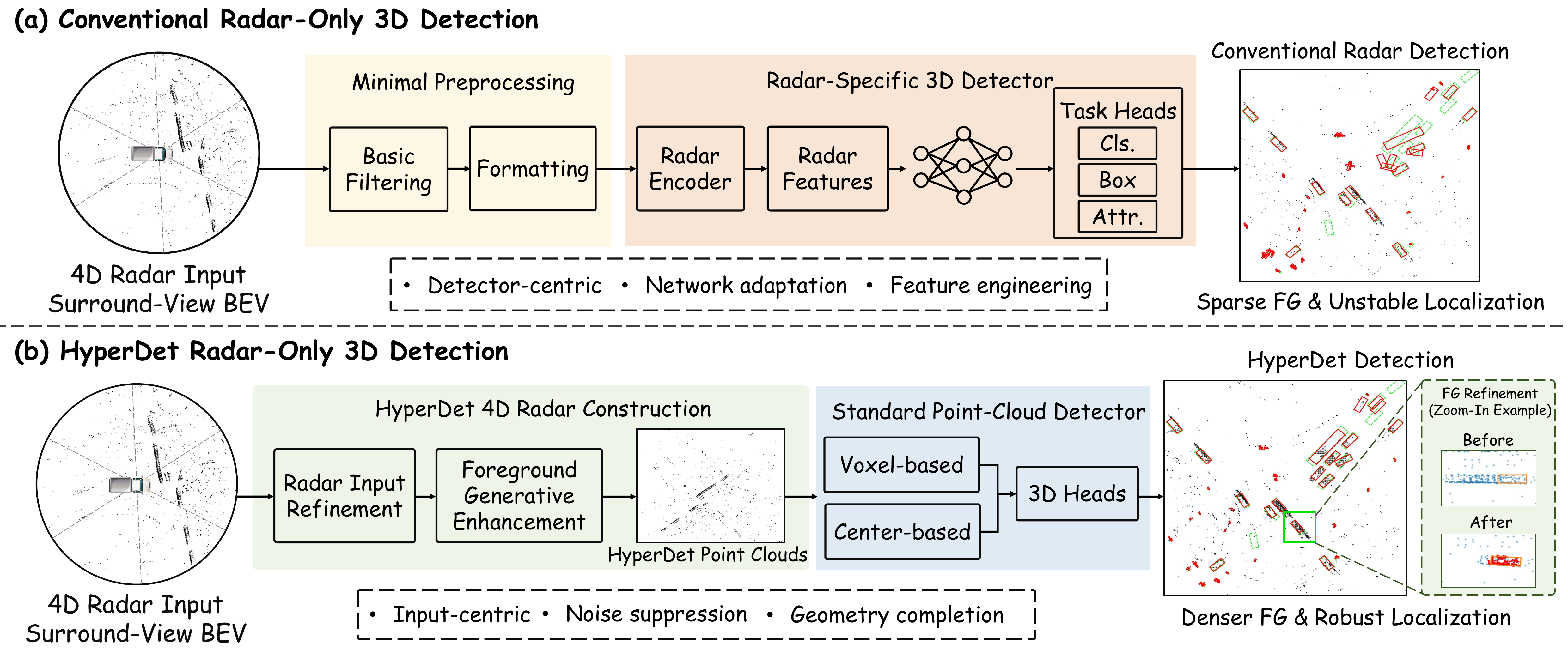}
    \caption{Comparison between conventional radar detection and \sysname. \textcolor{red}{Red} boxes denote predictions and \textcolor{green}{green} for GT.}
    \vspace{-2em}
    \label{fig:teaser}
\end{figure}

To address these challenges, we present \sysname, a detector-agnostic pipeline that constructs a task-aware \emph{hyper 4D radar point cloud} before detection by combining refined radar observations with completed foreground geometry. 
Instead of relying solely on detector-side adaptation to process weak radar evidence, \sysname improves the radar representation itself in a physically consistent manner. It first aggregates short-window surround-view observations to increase object-level point support, while cross-sensor validation suppresses weakly supported and geometrically inconsistent returns. 
Doppler-guided motion compensation further corrects the misalignment of moving-object returns, reducing temporal drift and trailing artifacts. Since the refined measurements still provide incomplete foreground geometry, we introduce foreground generative enhancement supervised by LiDAR-derived foreground geometry only during training. The completed foreground is combined with measured radar evidence rather than replacing the complete radar scene, enriching detection-relevant object structure while preserving measured background returns and radar-associated attributes. Finally, radar-aware object-level augmentation updates radial velocity after geometric relocation, maintaining its consistency with the new observation direction during detector training. At inference time, \sysname requires radar measurements alone and produces enhanced point clouds that can be directly consumed by standard 3D detectors without architectural modification.

We evaluate \sysname on MAN TruckScenes~\cite{fent2024man} and OmniHD-Scenes~\cite{zheng2026omnihd}, two public benchmarks featuring surround-view 4D radar sensing, using multiple standard 3D detectors~\cite{chen2023voxelnext,yin2021center} without modifying their architectures. We also compare against radar-only methods and matched temporal aggregation baselines under consistent input windows and detector settings. 
Beyond downstream detection performance, we assess the enhanced radar representation through LiDAR-referenced input analysis, object-level foreground recovery, motion alignment in observable dynamic scenes, generated-point reliability, component-wise ablations, and inference efficiency. Across datasets and detectors, \sysname consistently improves radar-only 3D detection under matched evaluation settings. These results demonstrate that strengthening radar evidence before detection is an effective complement to detector-centric design for robust robotic perception.

Our contributions are summarized as follows:

\begin{itemize}
    \setlength{\itemsep}{0pt}
    \setlength{\parskip}{0pt}
    \setlength{\parsep}{0pt}

    \item We present \sysname, a detector-agnostic pipeline that constructs task-aware hyper 4D radar point clouds by combining refined radar measurements with completed foreground geometry for radar-only 3D detection.

    \item We develop radar-aware input refinement that integrates surround-view spatio-temporal aggregation, cross-sensor validation, and Doppler-guided motion compensation to improve object-level geometric support and temporal coherence.

    \item We introduce foreground generative enhancement with training-time LiDAR-derived geometric supervision and radar-aware object-level augmentation, enriching foreground geometry while preserving attribute consistency.

    \item We validate \sysname on two public surround-view 4D radar benchmarks and standard detection backbones through matched radar baselines, component analyses, and representation-level evaluations.
\end{itemize}

\section{Related Work}

\noindent\textbf{LiDAR-Based 3D Object Detection.}
LiDAR provides accurate 3D measurements and has enabled mature point cloud detectors for autonomous driving and robotic perception~\cite{ghasemieh20223d,aung2024review}. Despite their geometric accuracy, LiDAR observations form discrete and unordered point sets that require suitable representations for effective feature extraction. Existing methods adopt voxel- or pillar-based grids, point- or graph-based formulations, and projected views such as bird's-eye view, balancing geometric fidelity, contextual aggregation, and computational efficiency~\cite{aung2024review}. Representative detectors include voxel-based methods such as VoxelNet~\cite{zhou2018voxelnet}, SECOND~\cite{yan2018second}, and VoxelNeXt~\cite{chen2023voxelnext}, pillar-based methods such as PointPillars~\cite{lang2019pointpillars}, and center-based BEV methods such as CenterPoint~\cite{yin2021center}. These mature point-cloud backbones motivate our detector-agnostic setting, where we improve the radar input rather than redesigning the detector.

\noindent\textbf{Radar-Only 3D Object Detection.}
Automotive radar has evolved from conventional sensing of range, azimuth, and radial velocity to modern 4D radar that additionally estimates elevation, providing richer 3D spatial observations~\cite{palffy2022VoD,paek2022k,zheng2022tj4dradset}. Early radar-only detectors largely reused LiDAR- or camera-oriented architectures, with limited treatment of radar-specific sparsity, noise, and measurement attributes~\cite{scheiner2021object}. With the emergence of public 4D radar benchmarks, radar-only perception methods have progressed along two complementary directions. Detector-oriented approaches introduce radar-specific feature encoding through velocity decomposition and pillar attention in RadarPillars~\cite{musiat2024radarpillars}, attention-based pillar and multi-scale features in PillarDAN~\cite{li2023pillarDan} and RMSA-Net~\cite{zhou2023rmsaNet}, graph representations in RadarGNN~\cite{fent2023radargnn}, and CFAR preprocessing with vertical encoding in RTNH+~\cite{kong2023rtnh+}. Input-oriented approaches improve radar measurements through filtering and temporal processing. Multi-frame accumulation increases point density but can disperse returns from independently moving objects~\cite{tan20223radardetection,palmer2023ego}, while DoppDrive reduces radial and tangential scatter through Doppler-guided point shifting and adaptive aggregation durations~\cite{haitman2025doppdrive}. Beyond these refinements, \sysname further enhances object-level geometry for detection while remaining compatible with standard point-cloud detectors.

\noindent\textbf{Radar Point Cloud Enhancement.}
Radar representations have been enhanced through noise filtering, ghost-target suppression, geometric reconstruction, and point cloud super-resolution. Learning-based methods identify unreliable returns or reconstruct denser spatial observations from sparse radar measurements~\cite{chamseddine2021ghost,cheng2022novel,geng2024dream}. Cross-modal methods further exploit LiDAR supervision to improve radar geometry or feature quality while retaining radar-only inference, as demonstrated by CORENet for radar denoising and discriminative feature learning~\cite{liu2025corenet}.
Generative models provide another approach to recovering missing radar geometry. Radar-Diffusion predicts dense LiDAR-like point clouds from sparse radar observations through cross-modal diffusion~\cite{zhang2024radardiffusion}, while diffusion-based radar super-resolution and R2LDM reconstruct denser geometry under LiDAR supervision~\cite{luan2024resolution,zheng2025r2ldm}. RadarGen instead synthesizes complete automotive radar point clouds, including geometry, RCS, and Doppler attributes, from multi-view images~\cite{borreda2025radargen}. 
These works demonstrate the potential of generative radar modeling for denoising, LiDAR-like reconstruction, super-resolution, and complete synthesis, while \sysname further enhances detection-relevant foreground geometry using measured radar evidence and preserves radar-only inference and native attributes.
\begin{figure*}[!t]
    \centering
    \includegraphics[width=0.94\textwidth]{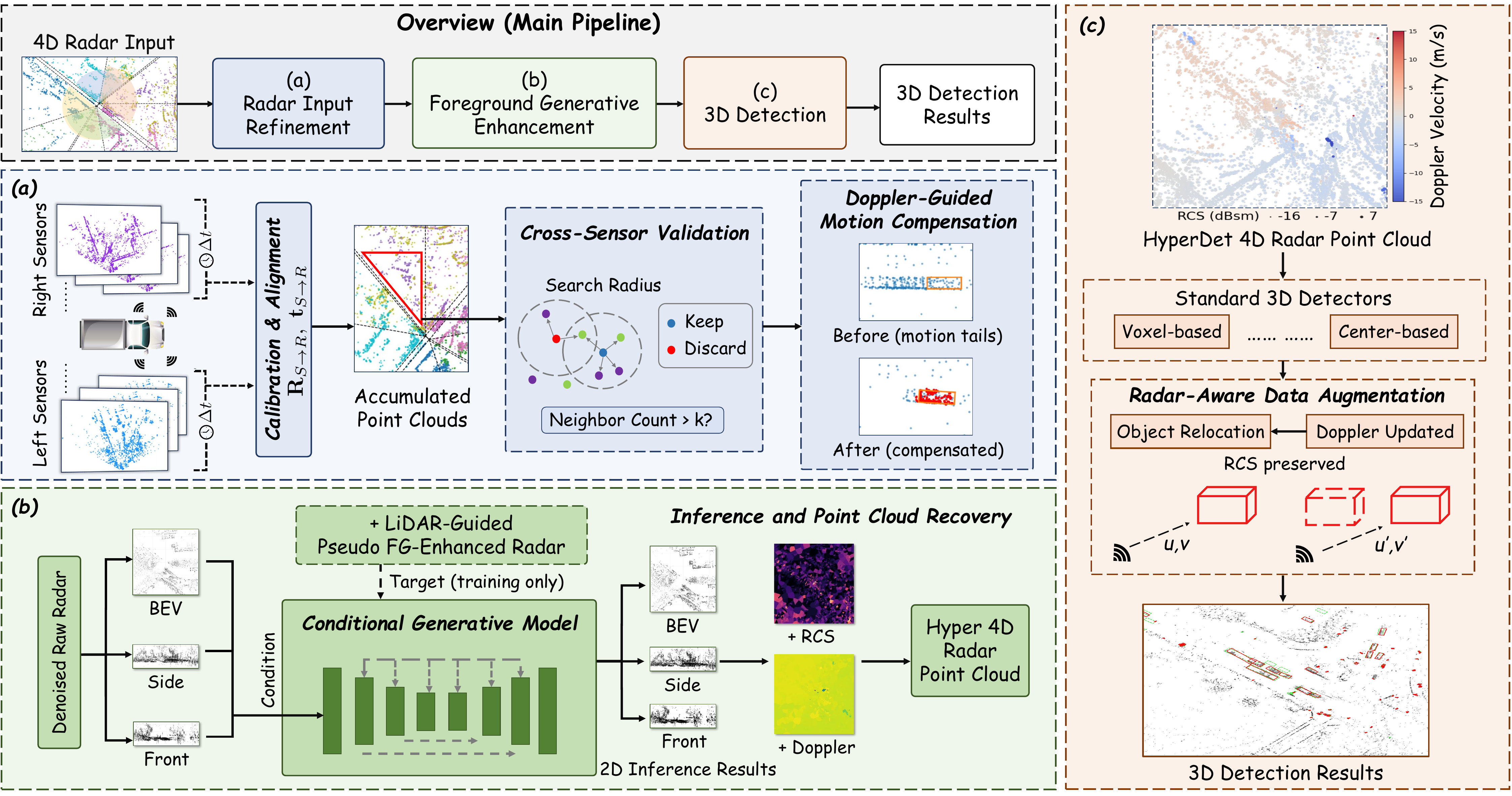}
    \caption{Overview of the proposed \sysname pipeline. Raw surround-view 4D radar sweeps are processed by (a) radar input refinement, (b) foreground generative enhancement with training-time LiDAR-guided supervision, and (c) 3D detection using standard point-cloud backbones.}
    \label{fig:pipeline}
    \vspace{-1.5em}
\end{figure*}

\section{Methodology}

\subsection{Task Definition}
\label{sec:task}

We study radar-only 3D object detection at inference time using short-window observations from surround-view 4D radars. Let
\(\mathcal{R}_{t}=\{R^{i}_{\tau}\mid i=1,\ldots,N,\;\tau=t-k,\ldots,t\}\)
denote radar sweeps collected by \(N\) surrounding radars over \(k+1\) frames ending at keyframe \(t\). Each sweep \(R^{i}_{\tau}\) consists of points
\(\mathbf{p}=[x,y,z,\sigma,v_d]\), where \((x,y,z)\) are Cartesian coordinates, \(\sigma\) denotes radar cross section, and \(v_d\) is the measured radial velocity. Each point is additionally associated with its acquisition time and source radar. Given \(\mathcal{R}_{t}\), \sysname constructs a task-aware hyper 4D radar point cloud \(\hat{\mathcal{P}}_t\) by combining refined radar observations with completed foreground geometry. LiDAR-derived geometry is used only to construct training supervision, while inference requires radar measurements alone.

\subsection{Overview}
\label{sec:overview}

As illustrated in Fig.~\ref{fig:pipeline}, \sysname constructs the hyper 4D radar point cloud through two enhancement stages. Radar input refinement first aligns and aggregates spatio-temporal observations, validates returns in overlapping sensor fields of view, and conditionally compensates moving-object misalignment, yielding an intermediate radar point cloud \(\tilde{\mathcal{P}}_t\). Foreground generative enhancement then uses training-time LiDAR supervision to complete detection-relevant object geometry. Candidates supported by consistent multi-view predictions and nearby radar observations are combined with \(\tilde{\mathcal{P}}_t\), preserving the retained measurements and their attributes. During detector training, radar-aware object-level augmentation updates radial velocity after geometric relocation. The resulting point cloud \(\hat{\mathcal{P}}_t\) is directly consumed by standard detectors without architectural modification.

\subsection{Radar Input Refinement}
\label{sec:input_refinement}

Given \(\mathcal{R}_t\), radar input refinement constructs \(\tilde{\mathcal{P}}_t\) through spatio-temporal alignment and accumulation, cross-sensor validation, and Doppler-guided motion compensation. This representation serves both as detector input and as the condition for subsequent foreground enhancement.

\noindent\textbf{Spatio-Temporal Alignment and Accumulation.}
We transform observations from radar \(i\) at time \(\tau\) into the reference frame of keyframe \(t\) using
\(\mathbf{T}_{t\leftarrow i,\tau}
=
\mathbf{T}^{\mathrm{ego}}_{t\leftarrow\tau}
\mathbf{T}^{\mathrm{ext}}_{\mathrm{ref}\leftarrow i}\),
where the two terms account for ego motion and sensor extrinsics, respectively. For
\(\mathbf{p}=[x,y,z,\sigma,v_d]\in R^i_\tau\), with
\(\bar{\mathbf{x}}=[x,y,z,1]^\top\), the accumulated point cloud is
\begin{equation}
\mathcal{P}^{\mathrm{acc}}_t
=
\bigcup_{i=1}^{N}
\bigcup_{\tau=t-k}^{t}
\left\{
\left[
\pi\left(
\mathbf{T}_{t\leftarrow i,\tau}\bar{\mathbf{x}}
\right)^\top,
\sigma,
v_d
\right]^\top
\;\middle|\;
\mathbf{p}\in R^i_\tau
\right\},
\end{equation}
where \(\pi(\cdot)\) extracts the transformed 3D coordinate. The sensor index and acquisition time are retained as metadata, while RCS and Doppler remain associated with their original measurements and source radars. This process increases point support and surround-view coverage but may also accumulate clutter and moving-object trails.

\noindent\textbf{Cross-Sensor Validation.}
Cross-sensor validation examines returns in overlapping radar fields of view. Let
\(\mathcal{P}^{\mathrm{acc}}_{t,i}\) denote accumulated returns from radar \(i\). For
\(\mathbf{q}_j\in\mathcal{P}^{\mathrm{acc}}_{t,i}\), the other radars that can observe it are
\begin{equation}
\mathcal{O}_j
=
\left\{
l\neq i
\;\middle|\;
\mathbf{q}_j\in\mathrm{FoV}_l
\right\}.
\end{equation}
Points with \(\mathcal{O}_j=\varnothing\) are left unchanged. In an overlapping region, a point is retained if
\begin{equation}
\begin{aligned}
&\exists\,l\in\mathcal{O}_j,\;
\exists\,\mathbf{q}_n\in\mathcal{P}^{\mathrm{acc}}_{t,l}
:\|\mathbf{q}_{j,xyz}-\mathbf{q}_{n,xyz}\|_2<\tau_d\\
&\hspace{3em}\lor\quad N_i(\mathbf{q}_j;r)> k,
\end{aligned}
\end{equation}
where \(\tau_d\) is the cross-radar support radius,
\(N_i(\mathbf{q}_j;r)\) counts returns from radar \(i\)
within radius \(r\) of \(\mathbf{q}_j\), and \(k\) is the
local neighbor-count threshold. Points failing both conditions are removed. Validation follows accumulation because individual sweeps are too sparse for reliable support. Returns from different timestamps can support one another. The retained points form \(\mathcal{P}^{\mathrm{val}}_t\).

\noindent\textbf{Doppler-Guided Motion Compensation.}
Ego-motion alignment corrects static returns but leaves independently moving objects misaligned. For point \(j\), let \(\mathbf{u}_j\) denote the line-of-sight direction from its source radar and \(\mathbf{v}_{s,j}\) the source-radar velocity at acquisition time, both expressed in the common frame. Under our Doppler convention, points with large static-scene residual
\(|r_j|\), where
\(r_j=v_{d,j}+\mathbf{u}_j^\top\mathbf{v}_{s,j}\),
are treated as dynamic candidates and grouped through spatial clustering. Assuming a shared planar velocity
\(\mathbf{v}_c=[v_x,v_y]^\top\)
for cluster \(c\), we estimate its motion from the radial constraints as
\begin{equation}
\hat{\mathbf{v}}_c
=
\arg\min_{\mathbf{v}\in\mathbb{R}^2}
\sum_{j\in c}
\left[
v_{d,j}
-
\mathbf{u}_{j,xy}^{\top}
\left(
\mathbf{v}-\mathbf{v}_{s,j,xy}
\right)
\right]^2.
\end{equation}
Because Doppler provides only radial constraints, compensation is applied only when the cluster contains sufficient evidence, its estimated motion is significant, and the least-squares system is well conditioned. Otherwise, the ego-motion-aligned points remain unchanged. For an accepted cluster, a point acquired at \(\tau_j\) is propagated to keyframe \(t\) as
\(\mathbf{x}^{\mathrm{ma}}_{j,xy}
=
\mathbf{x}^{\mathrm{acc}}_{j,xy}
+
(t-\tau_j)\hat{\mathbf{v}}_c\).
Its vertical coordinate, RCS, Doppler, timestamp, and source-radar metadata remain unchanged. The compensated and unchanged returns form \(\tilde{\mathcal{P}}_t\).

\subsection{Foreground Generative Enhancement}
\label{sec:foreground_enhancement}

Radar input refinement improves spatial support, return reliability, and temporal alignment, but \(\tilde{\mathcal{P}}_t\) still provides incomplete geometry around objects. On MAN TruckScenes~\cite{fent2024man}, fewer than \(2\%\) of radar returns lie inside annotated 3D boxes, compared with approximately \(10\%\) of LiDAR points. We therefore enhance detection-relevant foreground structure instead of reconstructing a globally dense LiDAR-like scene. The enhancement target adds LiDAR-derived object geometry to the measured radar scene, allowing the model to learn foreground completion while retaining the original radar background.

\noindent\textbf{LiDAR-Guided Foreground Supervision.}
LiDAR is used only during training to provide dense geometric supervision. We remove LiDAR ground points~\cite{lee2022patchwork++} and retain the remaining points inside annotated 3D boxes, forming a LiDAR-derived foreground set \(\mathcal{F}^{L}_t\). Let \(\pi_{xyz}(\tilde{\mathcal{P}}_t)\) denote the spatial coordinates of refined radar points. The target geometry is
\begin{equation}
\mathcal{G}^{\mathrm{tar}}_t
=
\pi_{xyz}(\tilde{\mathcal{P}}_t)
\cup
\mathcal{F}^{L}_t.
\end{equation}
Measured radar geometry remains unchanged, and additional supervision is confined to foreground object regions. LiDAR points provide target occupancy and geometry but are not physical radar measurements and receive no synthesized RCS or Doppler values.

\noindent\textbf{Multi-View Generative Training.}
We represent the radar condition and target geometry using three orthogonal occupancy views. Let
\(\mathcal{M}=\{\mathrm{BEV},\mathrm{front},\mathrm{side}\}\)
denote the \(xy\), \(xz\), and \(yz\) projections, respectively. For each \(m\in\mathcal{M}\), the conditioning and target maps are
\begin{equation}
\mathbf{c}^{m}
=
\mathrm{Proj}_{m}
\left(
\pi_{xyz}(\tilde{\mathcal{P}}_t)
\right),
\qquad
\mathbf{x}^{m}_0
=
\mathrm{Proj}_{m}
\left(
\mathcal{G}^{\mathrm{tar}}_t
\right).
\end{equation}
We train an independent conditional generator for each view using the same formulation without sharing parameters. For diffusion, the target is perturbed as
\(\mathbf{x}^{m}_{\sigma}
=
\mathbf{x}^{m}_0+\sigma\boldsymbol{\epsilon}\),
where
\(\boldsymbol{\epsilon}\sim\mathcal{N}(\mathbf{0},\mathbf{I})\).
The denoiser \(D_{\theta_m}\) learns to recover the clean target from its noisy version and radar condition:
\begin{equation}
\mathcal{L}^{m}_{\mathrm{diff}}
=
\mathbb{E}
\left[
w(\sigma)
\left\|
D_{\theta_m}
\left(
\mathbf{x}^{m}_{\sigma},
\sigma,
\mathbf{c}^{m}
\right)
-
\mathbf{x}^{m}_0
\right\|_2^2
\right],
\end{equation}
where \(w(\sigma)\) follows the adopted diffusion formulation~\cite{karras2022edm}. The iterative diffusion teacher is further distilled into a consistency student for single-step inference~\cite{song2023consistency}. We also investigate flow matching using the same view-specific conditions and targets~\cite{lipman2023flow}.

\noindent\textbf{Multi-View Recovery and Radar Support.}
At inference time, each generator receives only its radar condition \(\mathbf{c}^{m}\) and predicts an enhanced occupancy map for the corresponding view. Thresholding each prediction produces
\(\mathbf{B}^{m}=\mathbf{1}[\hat{\mathbf{x}}^{m}_0>\gamma]\),
where \(\gamma\) is the occupancy confidence threshold. A candidate
\(\mathbf{g}=[x,y,z]^\top\)
is retained only when all three projections agree:
\begin{equation}
\mathbf{B}^{\mathrm{BEV}}(x,y)
\mathbf{B}^{\mathrm{front}}(x,z)
\mathbf{B}^{\mathrm{side}}(y,z)
=
1.
\end{equation}
This consistency constraint removes candidates predicted in only one or two views. Each consistent candidate must also be supported by a nearby refined radar return:
\begin{equation}
\mathbf{p}^{*}(\mathbf{g})
=
\arg\min_{\mathbf{p}\in\tilde{\mathcal{P}}_t}
\left\|
\mathbf{g}-\mathbf{p}_{xyz}
\right\|_2,
\qquad
\left\|
\mathbf{g}-\mathbf{p}^{*}_{xyz}(\mathbf{g})
\right\|_2
\leq
\rho,
\end{equation}
where \(\rho\) is the radar-support radius. Candidates without measured radar support within this radius are discarded. Each retained candidate inherits RCS, Doppler, and source-radar metadata from \(\mathbf{p}^{*}(\mathbf{g})\). The generator therefore completes foreground geometry without synthesizing new physical radar attributes.

The retained candidates form \(\hat{\mathcal{F}}_t\), which is combined with the refined radar point cloud as
\(\hat{\mathcal{P}}_t
=
\tilde{\mathcal{P}}_t
\cup
\hat{\mathcal{F}}_t\).
The refined radar points and their associated attributes remain unchanged during generative enhancement, while generated points are added only when supported by multi-view agreement and nearby radar evidence.

\subsection{Radar-Aware Data Augmentation}
\label{sec:radar_augmentation}

Standard geometric augmentation can invalidate the radial-velocity attributes of radar points. In object-level copy-paste, relocating an object changes the line-of-sight direction between its returns and their source radars. Retaining the original Doppler values would therefore associate the relocated geometry with measurements acquired along different observation directions.

For copied point \(j\), let \(\mathbf{x}_j\) and \(\mathbf{x}'_j\) denote its original and relocated coordinates, and let \(\mathbf{s}_j\) denote the source-radar position. The corresponding line-of-sight directions are
\begin{equation}
\mathbf{u}_j
=
\frac{\mathbf{x}_j-\mathbf{s}_j}
{\|\mathbf{x}_j-\mathbf{s}_j\|},
\qquad
\mathbf{u}'_j
=
\frac{\mathbf{x}'_j-\mathbf{s}_j}
{\|\mathbf{x}'_j-\mathbf{s}_j\|}.
\end{equation}
Because a single return provides only radial velocity, we use the minimum-norm radial-motion approximation
\(v_{d,j}\mathbf{u}_j\)
and project it onto the relocated line of sight:
\begin{equation}
v'_{d,j}
=
\mathbf{u}'^{\top}_j
\left(
v_{d,j}\mathbf{u}_j
\right)
=
v_{d,j}\mathbf{u}'^{\top}_j\mathbf{u}_j.
\end{equation}
We preserve the measured RCS as
\(\sigma'_j=\sigma_j\),
rather than synthesizing its aspect-dependent change. This updates the radial-motion cue after relocation without estimating the unobserved full object velocity.
\section{Experiments}

% Figures and tables are declared before their first textual references.
% Their final positions should be checked in the compiled PDF.

\begin{table*}[t]
\centering
\caption{Radar-only 3D detection on MAN TruckScenes (mAP / NDS) and OmniHD-Scenes (mAP / ODS). Component comparisons use matched ten-sweep inputs. LiDAR-based rows are MAN-only reference upper bounds.}
\label{tab:detection_results}
\small
\renewcommand{\arraystretch}{1.05}
\setlength{\tabcolsep}{3pt}
\begin{tabular*}{\textwidth}{
@{\extracolsep{\fill}}
llccccc
@{}
}
\toprule
\textbf{Input / Method}
& \textbf{Detector}
& \textbf{Accum.}
& \textbf{Cross Val.}
& \textbf{Motion Comp.}
& \textbf{MAN}
& \textbf{OmniHD} \\
\midrule
\multicolumn{7}{l}{\textit{Matched temporal comparison and component analysis}} \\

Single-Sweep Radar
& VoxelNeXt
& & &
& 0.0619 / 0.1663
& 0.1403 / 0.2670 \\

Ten-Sweep Accumulation
& VoxelNeXt
& \(\checkmark\) & &
& 0.0958 / 0.1867
& 0.2700 / 0.2980 \\

+ Cross-Sensor Validation
& VoxelNeXt
& \(\checkmark\) & \(\checkmark\) &
& 0.1142 / 0.2027
& 0.3178 / 0.3851 \\

+ Motion Compensation
& VoxelNeXt
& \(\checkmark\) & & \(\checkmark\)
& 0.1092 / 0.2007
& 0.3100 / 0.3789 \\

Refined Radar
& VoxelNeXt
& \(\checkmark\) & \(\checkmark\) & \(\checkmark\)
& 0.1144 / 0.2084
& 0.3210 / 0.3864 \\

\midrule
\multicolumn{7}{l}{\textit{Radar-specific detector comparison}} \\

Ten-Sweep Radar
& RadarGNN
& \(\checkmark\) & &
& 0.1210 / 0.2054
& 0.3264 / 0.3872 \\

\midrule
\multicolumn{7}{l}{\textit{Foreground enhancement}} \\

Refined Radar + U-Net
& VoxelNeXt
& \(\checkmark\) & \(\checkmark\) & \(\checkmark\)
& 0.1203 / 0.2097
& 0.3205 / 0.3866 \\

Refined Radar + Student
& VoxelNeXt
& \(\checkmark\) & \(\checkmark\) & \(\checkmark\)
& 0.1276 / 0.2191
& 0.3315 / 0.3872 \\

Refined Radar + Diffusion
& VoxelNeXt
& \(\checkmark\) & \(\checkmark\) & \(\checkmark\)
& \textbf{0.1314 / 0.2217}
& 0.3347 / 0.3915 \\

Refined Radar + Flow Matching
& VoxelNeXt
& \(\checkmark\) & \(\checkmark\) & \(\checkmark\)
& 0.1285 / 0.2200
& \textbf{0.3472 / 0.3966} \\

\midrule
\multicolumn{7}{l}{\textit{Transfer to another standard detector}} \\

Single-Sweep Radar
& CenterPoint
& & &
& 0.0688 / 0.1724
& 0.1452 / 0.2635 \\

Ten-Sweep Accumulation
& CenterPoint
& \(\checkmark\) & &
& 0.0835 / 0.1816
& 0.2640 / 0.2710 \\

Refined Radar
& CenterPoint
& \(\checkmark\) & \(\checkmark\) & \(\checkmark\)
& 0.1221 / 0.2149
& 0.3362 / 0.3873 \\

Refined Radar + Diffusion
& CenterPoint
& \(\checkmark\) & \(\checkmark\) & \(\checkmark\)
& \textbf{0.1263 / 0.2188}
& \textbf{0.3544 / 0.4023} \\

Refined Radar + Flow Matching
& CenterPoint
& \(\checkmark\) & \(\checkmark\) & \(\checkmark\)
& 0.1239 / 0.2163
& 0.3426 / 0.3894 \\

\midrule
\multicolumn{7}{l}{\textit{Reference upper bounds on MAN TruckScenes}} \\

Single-Sweep LiDAR
& VoxelNeXt
& & &
& 0.2419 / 0.2755
& \multicolumn{1}{c}{MAN only} \\

Mixed R\&L
& VoxelNeXt
& & &
& 0.2637 / 0.3036
& \multicolumn{1}{c}{MAN only} \\

\bottomrule
\end{tabular*}
\vspace{-1em}
\end{table*}

\begin{table}[t]
\centering
\caption{Mean change in in-box radar returns per MAN TruckScenes sample after motion compensation. DoppDrive denotes its Doppler-driven aggregation in Algorithm~1~\cite{haitman2025doppdrive}.}
\label{tab:motion_analysis}
\small
\renewcommand{\arraystretch}{1.05}
\setlength{\tabcolsep}{4pt}
\begin{tabular*}{\linewidth}{
@{\extracolsep{\fill}}
lccc
@{}
}
\toprule
\textbf{Subset}
& \textbf{\(n\)}
& \textbf{Ours \(\Delta\)}
& \textbf{DoppDrive \(\Delta\)} \\
\midrule
All samples
& 1000 & \(\mathbf{+0.9}\) & \(-86\) \\
Highway
& 715 & \(\mathbf{+3.4}\) & \(-102\) \\
Ego speed \(>20\,\mathrm{m/s}\)
& 544 & \(\mathbf{+7.3}\) & \(-113\) \\
\bottomrule
\end{tabular*}
\vspace{-1em}
\end{table}

\begin{table}[t]
\centering
\caption{Foreground enhancement on MAN TruckScenes. Raw is the reported class-wise average number of points inside ground-truth boxes; Diff. Added and FM Added are additional points recovered by diffusion and flow matching. The final row sums the seven displayed class averages. T and S denote diffusion teacher and student in (b).}
\label{tab:foreground_analysis}
\small
\renewcommand{\arraystretch}{1.05}
\setlength{\tabcolsep}{4pt}

\begin{tabular*}{\linewidth}{
@{\extracolsep{\fill}}
lccc
@{}
}
\toprule
\multicolumn{4}{l}{\textit{(a) Foreground point enhancement}} \\
\midrule
\textbf{Class}
& \textbf{Raw}
& \textbf{Diff. Added}
& \textbf{FM Added} \\
\midrule
Car
& 6.13 & +11.71 & +16.19 \\
Truck
& 31.31 & +27.50 & +32.15 \\
Other Vehicle
& 98.70 & +63.24 & +72.15 \\
Trailer
& 78.72 & +95.60 & +87.45 \\
Traffic Sign
& 4.50 & +1.26 & +2.51 \\
Traffic Cone
& 0.92 & +0.27 & +0.80 \\
Pedestrian
& 4.85 & +1.70 & +2.20 \\
\midrule
\textbf{Sum of Listed Classes}
& \textbf{225.13}
& \textbf{+201.28}
& \textbf{+213.45} \\
\bottomrule
\end{tabular*}

\vspace{0.6em}

\begin{tabular*}{\linewidth}{
@{\extracolsep{\fill}}
lccc
@{}
}
\toprule
\multicolumn{4}{l}{\textit{(b) Geometric enhancement quality}} \\
\midrule
\textbf{Setting}
& \textbf{CD \(\downarrow\)}
& \textbf{HD \(\downarrow\)}
& \textbf{F-score \(\uparrow\)} \\
\midrule
LiDAR FG, T (512)
& 8.3224 & 32.7222 & 0.3399 \\
LiDAR FG + Radar, T (128)
& 2.0412 & 16.0697 & 0.7612 \\
LiDAR FG + Radar, S (512)
& 0.1689 & 7.6858 & 0.8140 \\
LiDAR FG + Radar, T (512)
& \textbf{0.0974}
& \textbf{4.4692}
& \textbf{0.9047} \\
\bottomrule
\end{tabular*}
\vspace{-1em}
\end{table}

\begin{table}[t]
\centering
\caption{Point recovery on MAN using VoxelNeXt.}
\label{tab:point_recovery}
\small
\renewcommand{\arraystretch}{1.05}
\setlength{\tabcolsep}{4pt}
\begin{tabular*}{\linewidth}{
@{\extracolsep{\fill}}
lccc
@{}
}
\toprule
\textbf{Setting}
& \textbf{Threshold}
& \textbf{mAP \(\uparrow\)}
& \textbf{NDS \(\uparrow\)} \\
\midrule
No Filling
& 60 & \textbf{0.1314} & \textbf{0.2217} \\
With Filling
& 60 & 0.1274 & 0.2113 \\
With Filling
& 200 & 0.1253 & 0.2097 \\
\bottomrule
\end{tabular*}
\vspace{-1em}
\end{table}

\begin{figure*}[t]
    \centering
    \begin{minipage}[t]{0.425\textwidth}
        \centering
        \includegraphics[width=\linewidth]{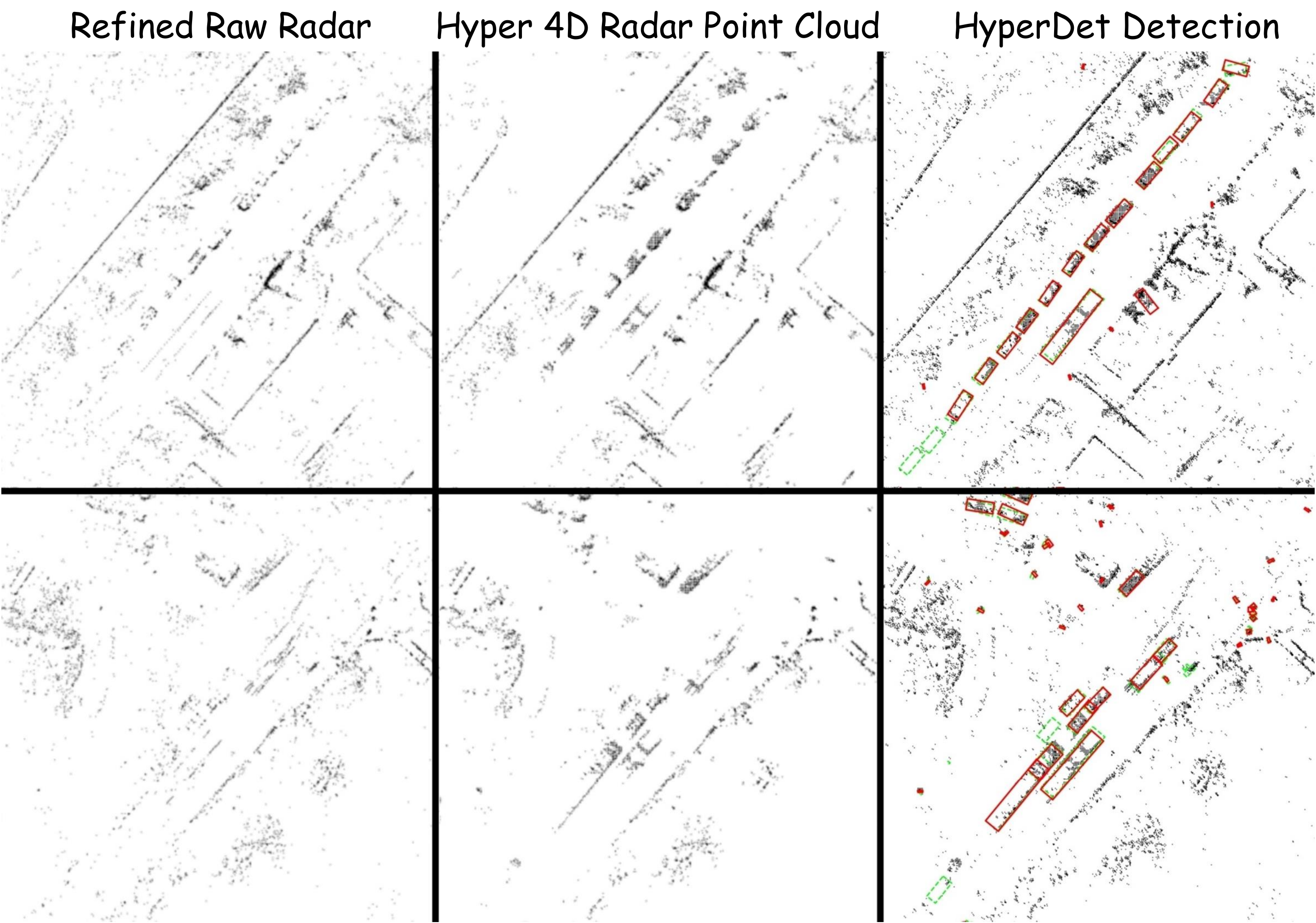}
        \vspace{-0.4em}
        \centerline{(a) Radar enhancement and 3D detection.}
    \end{minipage}
    \hfill
    \begin{minipage}[t]{0.565\textwidth}
        \centering
        \includegraphics[width=\linewidth]{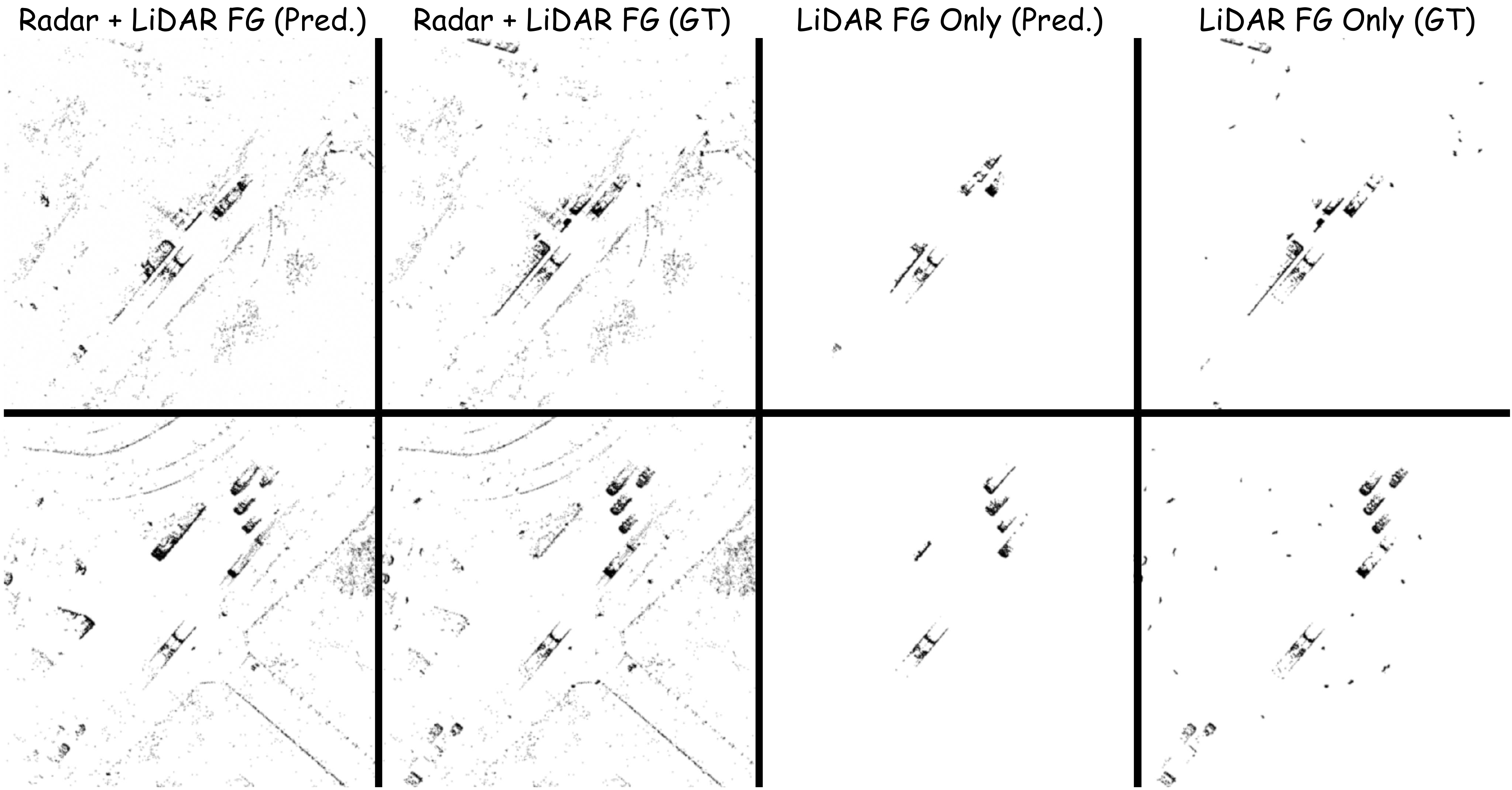}
        \vspace{-0.4em}
        \centerline{(b) Comparison of supervision targets.}
    \end{minipage}
    \caption{Qualitative results of \sysname. \textcolor{red}{Red} boxes denote predictions and \textcolor{green}{green} boxes denote ground truth.}
    \label{fig:qualitative_results}
    \vspace{-1.5em}
\end{figure*}

\begin{table}[t]
\centering
\caption{Component-wise computational cost}
\label{tab:efficiency}
\small
\renewcommand{\arraystretch}{1.05}
\setlength{\tabcolsep}{3pt}
\begin{tabular*}{\linewidth}{
@{\extracolsep{\fill}}
lccc
@{}
}
\toprule
\textbf{Component}
& \textbf{GFLOPs \(\downarrow\)}
& \textbf{Mem. (MB) \(\downarrow\)}
& \textbf{Latency (ms) \(\downarrow\)} \\
\midrule
ST Fusion
& \(\sim 0\) & 0.001 & 6.07 \\
Cross-Sensor Val.
& \(\sim 0\) & 0.001 & 32.74 \\
Motion Comp.
& \(\sim 0\) & 0.001 & 3.05 \\
\midrule
Flow Matching
& \(1.56\times10^{6}\)
& 2384.30
& 22647.69 \\
Diffusion Teacher
& \(2.49\times10^{4}\)
& 436.40
& 5008.78 \\
Consistency Student
& \(\mathbf{3.15\times10^{2}}\)
& \(\mathbf{428.40}\)
& \(\mathbf{68.97}\) \\
\midrule
VoxelNeXt
& 5.05 & 123.76 & 43.24 \\
\midrule
\textbf{Total, Flow}
& \(1.56\times10^{6}\)
& 2508.06
& 22732.79 \\
\textbf{Total, Teacher}
& \(2.49\times10^{4}\)
& 560.16
& 5093.88 \\
\textbf{Total, Student}
& \(\mathbf{3.20\times10^{2}}\)
& \(\mathbf{552.16}\)
& \(\mathbf{154.07}\) \\
\bottomrule
\end{tabular*}
\vspace{-1em}
\end{table}

\subsection{Experimental Setup}
\label{sec:setup}

\noindent\textbf{Datasets.}
We evaluate \sysname on two public datasets with synchronized surround-view measurements from six 4D radars and LiDAR: MAN TruckScenes~\cite{fent2024man} and OmniHD-Scenes~\cite{zheng2026omnihd}. LiDAR supplies training-time foreground supervision and a reference for geometric evaluation, while inference uses radar alone. MAN TruckScenes contains 747 annotated scenes and approximately 30,000 keyframes. We follow its official 70\%/10\%/20\% split, report validation results, and evaluate car, truck, other vehicle, trailer, traffic sign, traffic cone, and pedestrian within \(50\,\mathrm{m}\). Three colocated LiDARs provide surround-view foreground supervision. OmniHD-Scenes contains 200 annotated clips and 11,921 keyframes. We use its public 70\%/30\% training and evaluation split and evaluate car, large vehicle, rider, and pedestrian classes within the official region.

\noindent\textbf{Evaluation and Compared Methods.}
We evaluate \sysname from three perspectives.
\emph{(1) Radar input refinement quality.}
For cross-sensor validation, we report the unsupported-point ratio and radar-to-LiDAR F-score to assess return reliability and geometric agreement. For Doppler-guided motion compensation, FG / Box and FG Ratio measure the average number of foreground returns per annotated box and the proportion of returns in foreground regions. We also measure changes in in-box returns using fixed ground-truth boxes.
\emph{(2) Foreground generative enhancement quality.}
We compare recovered foreground geometry with the LiDAR-guided target using Chamfer Distance (CD), Hausdorff Distance (HD), and F-score.
\emph{(3) 3D detection performance.}
We report center-distance-based mAP with NDS on MAN TruckScenes and ODS on OmniHD-Scenes. Single-sweep radar provides a reference, while component comparisons use matched ten-sweep inputs. We compare the diffusion teacher, distilled student, and flow matching with a deterministic U-Net~\cite{ronneberger2015unet} to assess the benefit of generative modeling. We transfer the Algorithm~1 motion compensation module of DoppDrive~\cite{haitman2025doppdrive} to our accumulated input and evaluate in-box return changes in Table~\ref{tab:motion_analysis}, without reproducing its full detection pipeline. RadarGNN~\cite{fent2023radargnn} is evaluated on matched ten-sweep input in Table~\ref{tab:detection_results}.

\noindent\textbf{Implementation Details.}
We accumulate ten radar sweeps within \(0.5\,\mathrm{s}\) and align them to the keyframe coordinate system. On MAN TruckScenes, the common frame is \texttt{LIDAR\_LEFT}. Cross-sensor validation is applied after accumulation in overlapping radar fields of view. A return in an overlapping region is retained if it has support from another radar within \(10\,\mathrm{m}\) or sufficient local support from its originating radar. Returns outside overlapping regions are left unchanged. We restrict the effective fields of view of the two rear radars from \(120^\circ\) to \(100^\circ\) to account for truck-body self-occlusion.
Dynamic candidates are identified using a \(2.2\,\mathrm{m/s}\) static-scene Doppler residual. DBSCAN groups these candidates using a \(1.0\,\mathrm{m}\) neighborhood radius and a minimum of six points. Compensation is applied to clusters containing at least 20 points when the estimated planar speed exceeds \(6.0\,\mathrm{m/s}\) and the condition number of the velocity-estimation system is below 250. We limit displacement to \(15\,\mathrm{m}\). Points that do not meet these conditions are left unchanged.
The BEV, front, and side generators are trained independently using \(512\times512\) occupancy maps. We extract foreground candidates at an intensity threshold of 60 and retain those with a refined radar point within \(2.5\,\mathrm{m}\). Each retained candidate inherits RCS, Doppler, and source-radar metadata from its nearest measured point. We implement detection in OpenPCDet~\cite{openpcdet2020} using VoxelNeXt~\cite{chen2023voxelnext} and CenterPoint~\cite{yin2021center} without architectural modification. Unless otherwise stated, detectors are trained for 20 epochs with matched optimization schedules and augmentation settings.

\subsection{Quantitative Results}

\noindent\textbf{Main 3D Detection Results.}
Table~\ref{tab:detection_results} separates temporal accumulation from the contributions of input refinement and foreground generation. On MAN TruckScenes with VoxelNeXt, ten-sweep accumulation achieves \(0.0958\) mAP. Cross-sensor validation and motion compensation each improve this matched baseline, and their combination reaches \(0.1144\) mAP. Diffusion enhancement further raises mAP to \(0.1314\) and NDS to \(0.2217\). Because all component configurations use the same ten sweeps, these gains reflect changes to the radar input rather than a longer accumulation window.
The complete diffusion pipeline also exceeds RadarGNN on the matched ten-sweep MAN input. A deterministic U-Net improves refined radar but remains below all three generative variants on MAN TruckScenes, showing that foreground prediction alone does not account for the full gain from generative modeling. The single-step student retains most of the diffusion teacher's improvement. Similar improvement over matched accumulation is observed on OmniHD-Scenes and with CenterPoint. Flow matching achieves the highest VoxelNeXt mAP and ODS on OmniHD-Scenes.
As additional references on MAN TruckScenes, we evaluate single-sweep LiDAR and Mixed R\&L to contextualize the remaining radar-to-LiDAR gap. For Mixed R\&L, LiDAR foreground points are assigned attributes from nearby radar measurements to form a detector-compatible input.

\noindent\textbf{Radar Input Refinement Quality.}
Matched detection results establish the benefit of input refinement but not how the representation changes. On MAN TruckScenes, cross-sensor validation reduces the unsupported-point ratio from \(0.4349\) to \(0.4118\) and raises the radar-to-LiDAR F-score from \(0.6648\) to \(0.6781\). These changes show improved return reliability while preserving support from other radars or local density in the source radar. Motion compensation reduces temporal trails after ego-motion alignment, increasing foreground returns per box from \(12.1246\) to \(12.4612\) and the foreground-return ratio from \(0.0174\) to \(0.0199\). Because these measures do not directly establish alignment accuracy, we additionally evaluate in-box return changes.
To isolate dynamic-motion handling, we reproduce the detector-independent Doppler-driven aggregation defined in Algorithm~1 of DoppDrive~\cite{haitman2025doppdrive} and apply it to the same ten-sweep histories with fixed ground-truth boxes. Table~\ref{tab:motion_analysis} shows that our module gains \(0.9\) in-box returns overall, \(3.4\) on highways, and \(7.3\) above \(20\,\mathrm{m/s}\). DoppDrive instead loses returns because its pointwise duration gate discards older measurements more aggressively as speed rises, reflecting its adaptive aggregation design rather than coordinate-alignment failure. By contrast, our conservative cluster-level method compensates only clusters with reliable planar-motion estimates and leaves all other measurements unchanged. This module-level comparison shows that our conservative cluster-level compensation better preserves in-box evidence in our pipeline than DoppDrive's pointwise Doppler propagation with adaptive duration gating.

\noindent\textbf{Foreground Generative Enhancement Quality.}
Input refinement improves measured radar evidence but cannot recover unobserved object geometry. Table~\ref{tab:foreground_analysis}(a) shows that diffusion and flow matching add points inside annotated boxes across all seven MAN TruckScenes classes. The displayed class averages sum to \(225.13\) raw points, with \(201.28\) added by diffusion and \(213.45\) by flow matching. These additions correspond to \(89.41\%\) and \(94.81\%\) of the raw sum. The final row sums class averages rather than reporting a separately computed dataset-wide mean.

Additional points are useful only when the recovered structure is geometrically coherent. Table~\ref{tab:foreground_analysis}(b) compares generated geometry with the LiDAR-guided target. LiDAR-foreground-only supervision gives a CD of \(8.3224\) and F-score of \(0.3399\). Including measured radar geometry in the target improves agreement. At resolution 512, the diffusion teacher reaches a CD of \(0.0974\), HD of \(4.4692\), and F-score of \(0.9047\). The student has lower geometric fidelity but retains most of the teacher's detection improvement. These measurements assess agreement with the supervision target. The separate radar-support constraint limits how far newly recovered points can lie from measured evidence.

\subsection{Ablation Studies}

\noindent\textbf{Point Recovery and Parameter Choices.}
Point recovery determines which generated candidates enter the enhanced radar cloud. Table~\ref{tab:point_recovery} compares threshold-based recovery with heuristic filling. Recovery without filling achieves \(0.1314\) mAP and \(0.2217\) NDS, outperforming both filling configurations. Adding more generated points therefore does not necessarily improve detection, motivating our use of geometrically consistent, radar-supported candidates. We select refinement parameters by downstream detection performance rather than visual sparsity. Cross-sensor validation radii of \(1\), \(2\), \(5\), and \(10\,\mathrm{m}\) were tested with validation as the sole refinement component. The \(10\,\mathrm{m}\) radius achieves the highest tested result of \(0.1142\) mAP and \(0.2027\) NDS. Among occupancy thresholds of 30, 60, 180, and 200, threshold 60 gives the highest tested result without point filling.

\noindent\textbf{Radar-Aware Object Augmentation.}
Object relocation changes the direction from a copied radar return to its source sensor. On ten-sweep raw MAN TruckScenes radar, conventional LiDAR-style copy-paste retains the original Doppler measurement and yields an mAVE of \(2.4891\). Radar-aware augmentation reprojects Doppler after relocation and reduces mAVE to \(0.8556\) under the same radar input. For context, single-sweep LiDAR with conventional LiDAR-style augmentation yields an mAVE of \(1.4007\). The matched radar comparison demonstrates the value of preserving radial-velocity consistency during relocation, and the LiDAR configuration is a separate sensor-input reference.

\subsection{Qualitative Results}

Fig.~\ref{fig:qualitative_results} illustrates the two enhancement stages of \sysname. The left part compares refined radar, the enhanced point cloud, and radar-only detections. Input refinement reduces motion trails, while foreground generation adds structure for estimating object extent and orientation.
The right part compares supervision targets. LiDAR-foreground-only supervision produces less localized, surface-like structures. Including measured radar observations yields more coherent foreground recovery, consistent with Table~\ref{tab:foreground_analysis}(b). Distant or small objects remain challenging with sparse returns, and strong static reflections may cause false positives. \sysname enhances observed evidence but cannot reliably recover targets absent from radar input.

\subsection{Runtime Efficiency}

Table~\ref{tab:efficiency} reports costs measured on a system equipped with an RTX~4090 GPU. Radar input refinement is lightweight, taking \(41.86\,\mathrm{ms}\) across fusion, cross-sensor validation, and motion compensation. The VoxelNeXt detector adds \(43.24\,\mathrm{ms}\). Foreground generation is the main bottleneck. The 40-step diffusion teacher takes \(5008.78\,\mathrm{ms}\), whereas consistency distillation reduces generation to \(68.97\,\mathrm{ms}\) and the complete pipeline to \(154.07\,\mathrm{ms}\) while retaining most of the teacher's detection gain. 
\section{Conclusion}

We presented \sysname, a detector-agnostic pipeline that constructs task-aware hyper 4D radar point clouds for radar-only 3D object detection. Spatio-temporal accumulation strengthens object evidence, cross-sensor validation suppresses unsupported returns, and Doppler-guided compensation reduces observable motion trails. Foreground generation recovers additional geometry under training-time LiDAR guidance while preserving measured radar points and attributes. Experiments on two surround-view datasets show gains over matched ten-sweep accumulation with standard 3D detectors, while input-quality and foreground analyses support improved return reliability, foreground support, and geometric recovery. These findings demonstrate the value of radar input enhancement alongside detector-centric design.

\noindent\textbf{Limitations and Future Work.}
The effectiveness of \sysname depends on the availability and quality of measured radar returns. Motion compensation requires reliable planar-velocity estimates, and distant or small objects, severe multipath, and targets with little radar evidence remain challenging. Fixed validation, motion, and recovery criteria may also require calibration for radar systems with different sensing characteristics. Generated points inherit nearby measured RCS and Doppler for detector compatibility rather than representing new physical measurements at their recovered locations. Future work will investigate adaptive refinement across radar configurations, uncertainty-aware foreground generation, and physically grounded radar-attribute estimation while preserving radar-only inference and detector compatibility.

%%%%%%%%%%%%%%%%%%%%%%%%%%%%%%%%%%%%%%%%%%%%%%%%%%%%%%%%%%%%%%%%%%%%%%%%%%%%%%%%

\bibliographystyle{IEEEtran}
\bibliography{ref}

@article{Karangwa2023review,
  title   = {Vehicle Detection for Autonomous Driving: A Review of Algorithms and Datasets},
  author  = {Karangwa, J. and Liu, J. and Zeng, Z.},
  journal = {T-ITS},
  year    = {2023}
}

@article{venon2022survey,
  title   = {Millimeter Wave FMCW Radars for Perception, Recognition and Localization in Automotive Applications: A Survey},
  author  = {Venon, A. and Dupuis, Y. and Vasseur, P. and Merriaux, P.},
  journal = {T-IV},
  year    = {2022}
}

@INPROCEEDINGS{Resilient24Chong,
  author={Zhang, Chong and Jin, Jin and Frey, Jonas and Rudin, Nikita and Mattamala, Matías and Cadena, Cesar and Hutter, Marco},
  booktitle={2024 IEEE International Conference on Robotics and Automation (ICRA)}, 
  title={{Resilient Legged Local Navigation}: Learning to Traverse with Compromised Perception End-to-End}, 
  year={2024},
  volume={},
  number={},
  pages={34-41},
  doi={10.1109/ICRA57147.2024.10611254}}

@inproceedings{ding2023hidden,
  title     = {Hidden Gems: 4D Radar Scene Flow Learning Using Cross-Modal Supervision},
  author    = {Ding, F. and Palffy, A. and Gavrila, D. M. and Lu, C. X.},
  booktitle = {CVPR},
  year      = {2023}
}

@article{ding2024radarocc,
  title   = {RadarOcc: Robust 3D Occupancy Prediction with 4D Imaging Radar},
  author  = {Ding, F. and Wen, X. and Zhu, Y. and Li, Y. and Lu, C. X.},
  journal = {NeurIPS},
  year    = {2024}
}

@article{palffy2022VoD,
  title   = {Multi-Class Road User Detection with 3+1D Radar in the View-of-Delft Dataset},
  author  = {Palffy, A. and Pool, E. and Baratam, S. and Kooij, J. F. P. and Gavrila, D. M.},
  journal = {RA-L},
  year    = {2022}
}

@article{jiang2024resolution,
  title   = {4D High-Resolution Imagery of Point Clouds for Automotive mmWave Radar},
  author  = {Jiang, M. and others},
  journal = {T-ITS},
  year    = {2024}
}

@inproceedings{zhou2023rmsaNet,
  title     = {RMSA-Net: A 4D Radar-Based Multi-Scale Attention Network for 3D Object Detection},
  author    = {Zhou, Y. and Hao, J. and Zhu, K.},
  booktitle = {ISCSIC},
  year      = {2023}
}

@inproceedings{musiat2024radarpillars,
  title     = {RadarPillars: Efficient Object Detection from 4D Radar Point Clouds},
  author    = {Musiat, A. and Reichardt, L. and Schulze, M. and Wasenm{\"u}ller, O.},
  booktitle = {ITSC},
  year      = {2024}
}

@inproceedings{li2023pillarDan,
  title     = {PillarDAN: Pillar-Based Dual Attention Network for 3D Object Detection with 4D Radar},
  author    = {Li, J. and Yang, L. and Chen, Y. and Yang, Y. and Jin, Y. and Akiyama, K.},
  booktitle = {ITSC},
  year      = {2023}
}

@article{tan20223radardetection,
  title   = {3D Object Detection for Multiframe 4D Automotive Millimeter-Wave Radar Point Clouds},
  author  = {Tan, B. and others},
  journal = {SensorsJ},
  year    = {2022}
}

@inproceedings{palmer2023ego,
  title     = {Ego-Motion Estimation and Dynamic Motion Separation from 3D Point Clouds for Accumulation and 3D Object Detection},
  author    = {Palmer, P. and Krueger, M. and Altendorfer, R. and Bertram, T.},
  booktitle = {AmE},
  year      = {2023}
}

@article{fent2024man,
  title   = {MAN TruckScenes: A Multimodal Dataset for Autonomous Trucking in Diverse Conditions},
  author  = {Fent, F. S. and others},
  journal = {NeurIPS},
  year    = {2024}
}

@article{zheng2026omnihd,
  title   = {OmniHD-Scenes: A Next-Generation Multimodal Dataset for Autonomous Driving},
  author  = {Zheng, L. and others},
  journal = {TPAMI},
  year    = {2026}
}

@inproceedings{chen2023voxelnext,
  title     = {VoxelNeXt: Fully Sparse VoxelNet for 3D Object Detection and Tracking},
  author    = {Chen, Y. and Liu, J. and Zhang, X. and Qi, X. and Jia, J.},
  booktitle = {CVPR},
  year      = {2023}
}

@inproceedings{yin2021center,
  title     = {Center-Based 3D Object Detection and Tracking},
  author    = {Yin, T. and Zhou, X. and Kr{\"a}henb{\"u}hl, P.},
  booktitle = {CVPR},
  year      = {2021}
}

@article{ghasemieh20223d,
  title   = {3D Object Detection for Autonomous Driving: Methods, Models, Sensors, Data, and Challenges},
  author  = {Ghasemieh, A. and Kashef, R.},
  journal = {Technologies},
  year    = {2022}
}

@article{aung2024review,
  title   = {A Review of LiDAR-Based 3D Object Detection via Deep Learning Approaches towards Robust Connected and Autonomous Vehicles},
  author  = {Aung, N. H. H. and Sangwongngam, P. and Jintamethasawat, R. and Shah, S. and Wuttisittikulkij, L.},
  journal = {T-IV},
  year    = {2024}
}

@inproceedings{zhou2018voxelnet,
  title     = {VoxelNet: End-to-End Learning for Point Cloud Based 3D Object Detection},
  author    = {Zhou, Y. and Tuzel, O.},
  booktitle = {CVPR},
  year      = {2018}
}

@article{yan2018second,
  title   = {SECOND: Sparsely Embedded Convolutional Detection},
  author  = {Yan, Y. and Mao, Y. and Li, B.},
  journal = {Sensors},
  year    = {2018}
}

@inproceedings{lang2019pointpillars,
  title     = {PointPillars: Fast Encoders for Object Detection from Point Clouds},
  author    = {Lang, A. H. and Vora, S. and Caesar, H. and Zhou, L. and Yang, J. and Beijbom, O.},
  booktitle = {CVPR},
  year      = {2019}
}

@article{paek2022k,
  title   = {K-Radar: 4D Radar Object Detection for Autonomous Driving in Various Weather Conditions},
  author  = {Paek, D.-H. and Kong, S.-H. and Wijaya, K. T.},
  journal = {NeurIPS},
  year    = {2022}
}

@inproceedings{zheng2022tj4dradset,
  title     = {TJ4DRadSet: A 4D Radar Dataset for Autonomous Driving},
  author    = {Zheng, L. and others},
  booktitle = {ITSC},
  year      = {2022}
}

@article{scheiner2021object,
  title   = {Object Detection for Automotive Radar Point Clouds: A Comparison},
  author  = {Scheiner, N. and Kraus, F. and Appenrodt, N. and Dickmann, J. and Sick, B.},
  journal = {Automotive and Engine Technology},
  year    = {2021}
}

@inproceedings{fent2023radargnn,
  title     = {RadarGNN: Transformation-Invariant Graph Neural Network for Radar-Based Perception},
  author    = {Fent, F. and Bauerschmidt, P. and Lienkamp, M.},
  booktitle = {CVPR},
  year      = {2023}
}

@article{kong2023rtnh+,
  title   = {RTNH+: Enhanced 4D Radar Object Detection Network Using Combined CFAR-Based Two-Level Preprocessing and Vertical Encoding},
  author  = {Kong, S.-H. and Paek, D.-H. and Cho, S.},
  journal = {arXiv},
  year    = {2023}
}

@inproceedings{haitman2025doppdrive,
  title     = {{DoppDrive}: Doppler-Driven Temporal Aggregation for Improved Radar Object Detection},
  author    = {Haitman, Y. and Bialer, O.},
  booktitle = {ICCV},
  year      = {2025}
}

@inproceedings{chamseddine2021ghost,
  title     = {Ghost Target Detection in 3D Radar Data Using Point-Cloud-Based Deep Neural Networks},
  author    = {Chamseddine, M. and Rambach, J. and Stricker, D. and Wasenm{\"u}ller, O.},
  booktitle = {ICPR},
  year      = {2021}
}

@article{cheng2022novel,
  title   = {A Novel Radar Point Cloud Generation Method for Robot Environment Perception},
  author  = {Cheng, Y. and Su, J. and Jiang, M. and Liu, Y.},
  journal = {T-RO},
  year    = {2022}
}

@article{geng2024dream,
  title   = {Dream-PCD: Deep Reconstruction and Enhancement of mmWave Radar Point Clouds},
  author  = {Geng, R. and others},
  journal = {TIP},
  year    = {2024}
}

@inproceedings{liu2025corenet,
  title     = {{CORENet}: Cross-Modal 4D Radar Denoising Network with {LiDAR} Supervision for Autonomous Driving},
  author    = {Liu, F. and Mei, J. and Mao, F. and Min, C. and Xing, Y. and Hu, Y.},
  booktitle = {IROS},
  year      = {2025}
}

@article{zhang2024radardiffusion,
  title   = {Towards Dense and Accurate Radar Perception via Efficient Cross-Modal Diffusion Model},
  author  = {Zhang, R. and Xue, D. and Wang, Y. and Geng, R. and Gao, F.},
  journal = {RA-L},
  year    = {2024}
}

@inproceedings{luan2024resolution,
  title     = {Diffusion-Based Point Cloud Super-Resolution for mmWave Radar Data},
  author    = {Luan, K. and Shi, C. and Wang, N. and Cheng, Y. and Lu, H. and Chen, X.},
  booktitle = {ICRA},
  year      = {2024}
}

@inproceedings{zheng2025r2ldm,
  title     = {{R2LDM}: An Efficient 4D Radar Super-Resolution Framework Leveraging Diffusion Model},
  author    = {Zheng, B. and others},
  booktitle = {IROS},
  year      = {2025}
}

@article{borreda2025radargen,
  title   = {RadarGen: Automotive Radar Point Cloud Generation from Cameras},
  author  = {Borreda, T. and Ding, F. and Fidler, S. and Huang, S. and Litany, O.},
  journal = {arXiv},
  year    = {2025}
}

@inproceedings{lee2022patchwork++,
  title     = {Patchwork++: Fast and Robust Ground Segmentation Solving Partial Under-Segmentation Using 3D Point Clouds},
  author    = {Lee, S. and Lim, H. and Myung, H.},
  booktitle = {IROS},
  year      = {2022}
}

@inproceedings{karras2022edm,
  title     = {Elucidating the Design Space of Diffusion-Based Generative Models},
  author    = {Karras, T. and Aittala, M. and Aila, T. and Laine, S.},
  booktitle = {NeurIPS},
  year      = {2022}
}

@inproceedings{song2023consistency,
  title     = {Consistency Models},
  author    = {Song, Y. and Dhariwal, P. and Chen, M. and Sutskever, I.},
  booktitle = {ICML},
  year      = {2023}
}

@inproceedings{lipman2023flow,
  title     = {Flow Matching for Generative Modeling},
  author    = {Lipman, Y. and Chen, R. T. Q. and Ben-Hamu, H. and Nickel, M. and Le, M.},
  booktitle = {ICLR},
  year      = {2023}
}

@inproceedings{ronneberger2015unet,
  title     = {U-Net: Convolutional Networks for Biomedical Image Segmentation},
  author    = {Ronneberger, O. and Fischer, P. and Brox, T.},
  booktitle = {MICCAI},
  year      = {2015}
}

@misc{openpcdet2020,
  title        = {OpenPCDet: An Open-Source Toolbox for 3D Object Detection from Point Clouds},
  author       = {{OpenPCDet Development Team}},
  howpublished = {\url{https://github.com/open-mmlab/OpenPCDet}},
  year         = {2020}
}

\end{document}